\documentclass{article}
\usepackage{spconf,amsmath,graphicx}
\usepackage{subfigure}

\title{On mini-batch training with varying length time series}
\name{Brian Kenji Iwana
\thanks{This work was partially supported by MEXT-Japan
(Grant No. J21K17808) and R3QR Program (Qdai-jump Research Program) 01252.}}
\address{Department of Advanced Information Technology \\ Kyushu University, Fukuoka, Japan}

\begin{document}
\ninept 
\maketitle
\begin{abstract}
In real-world time series recognition applications, it is possible to have data with varying length patterns. However, when using artificial neural networks (ANN), it is standard practice to use fixed-sized mini-batches. To do this, time series data with varying lengths are typically normalized so that all the patterns are the same length. Normally, this is done using zero padding or truncation without much consideration. We propose a novel method of normalizing the lengths of the time series in a dataset by exploiting the dynamic matching ability of Dynamic Time Warping (DTW). In this way, the time series lengths in a dataset can be set to a fixed size while maintaining features typical to the dataset. In the experiments, all 11 datasets with varying length time series from the 2018 UCR Time Series Archive are used. We evaluate the proposed method by comparing it with 18 other length normalization methods on a Convolutional Neural Network (CNN), a Long-Short Term Memory network (LSTM), and a Bidirectional LSTM (BLSTM). The code is publicly available at https://github.com/uchidalab/vary\_length\_time\_series.
\end{abstract}
\section{Introduction}
\label{sec:intro}

One challenge with tackling real-world time series data is processing time series of different lengths. 
Time series within a dataset can vary in length for many reasons. 
Some of the reasons include variations in the sampling, frequencies, observations relative to the signal, and variations in the start or end relative to other patterns~\cite{tan2019time}.

Traditionally, this issue can been addressed using distance-based methods~\cite{ding2008querying,Leng_2008,Jalalian_2013,tan2019time}. 
These methods use the distances between patterns for classification. 
Dynamic Time Warping (DTW)~\cite{sakoe1978dynamic}, in particular, is a popular distance measure able to overcome temporal distortions, such as variable lengths.  
Alternatively, other classic methods addressing varying length time series include wavelet-based methods~\cite{Kahveci} and feature extraction methods~\cite{Chandrakala_2010}.

Artificial Neural Networks (ANN), such as Convolutional Neural Networks (CNN)~\cite{Lecun_1998} and Recurrent Neural Networks (RNN)~\cite{rumelhart1988learning}, have had wide-reaching successes in time series classification~\cite{Wang_2017,bai2018empirical}. 
Furthermore, it has become common practice to train ANNs using mini-batch training. 
Much like bootstrapping, mini-batches are small subsets of the dataset used to train an ANN model during each gradient descent step. 
The advantage of using mini-batches is that it combines the stability of traditional full batch methods with the speed of stochastic methods~\cite{ruder2016overview}.

\begin{figure}[t]
    \centering
    \subfigure[Original]{
    \label{subfig:original}
    \includegraphics[width=0.30\columnwidth,trim={1.1cm 0.5cm 0.4cm 0.3cm},clip]{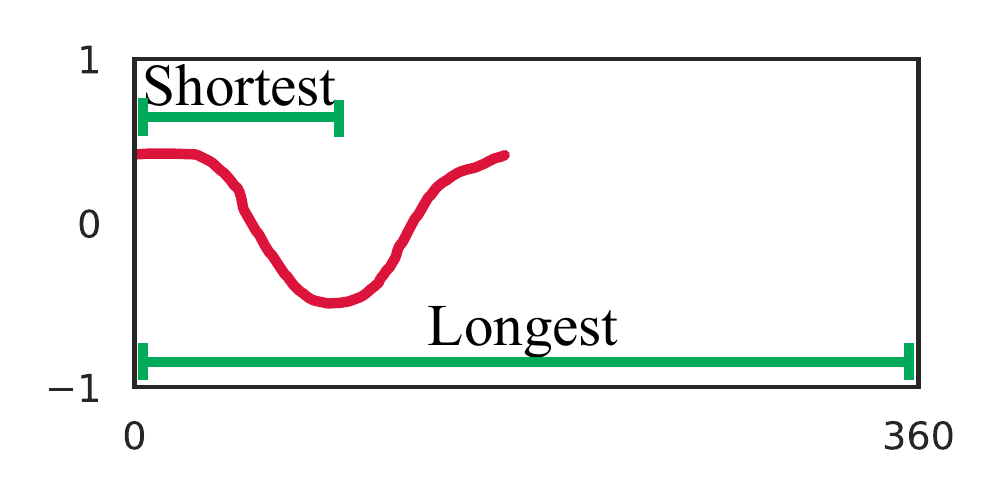}
    }
        \subfigure[Zero Pad (Pre)]{
    \label{subfig:zeropad}
    \includegraphics[width=0.30\columnwidth,trim={1.1cm 0.5cm 0.4cm 0.3cm},clip]{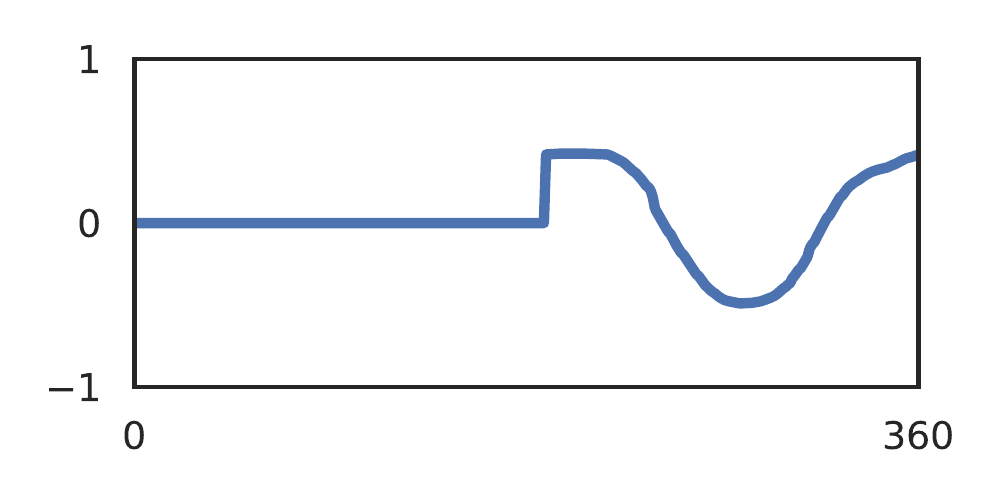}
    }
    \subfigure[Truncate (Post)]{
    \label{subfig:truncate}
    \includegraphics[width=0.30\columnwidth,trim={1.1cm 0.5cm 0.4cm 0.3cm},clip]{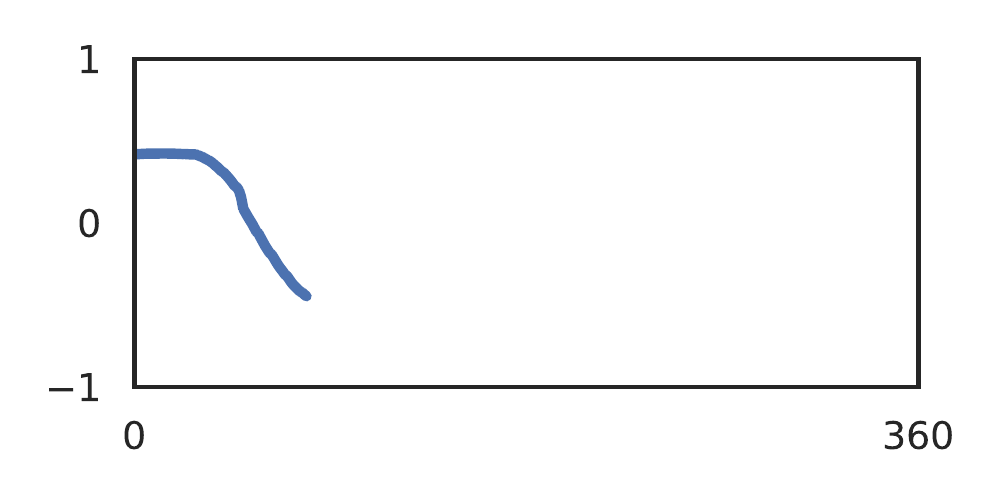}
    }
    \subfigure[Resampling]{
    \label{subfig:interpolated}
    \includegraphics[width=0.30\columnwidth,trim={1.1cm 0.5cm 0.4cm 0.3cm},clip]{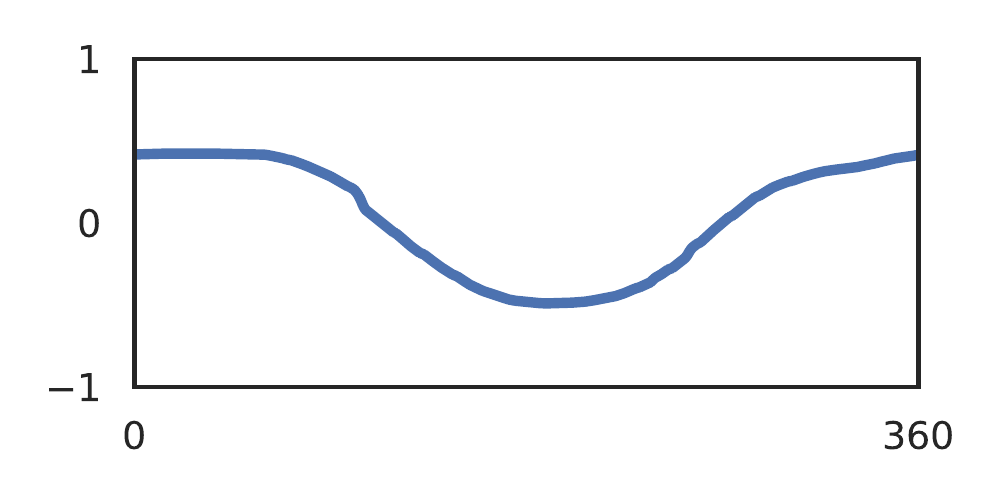}
    }
    \subfigure[Proposed]{
    \label{subfig:proposed}
    \includegraphics[width=0.30\columnwidth,trim={1.1cm 0.5cm 0.4cm 0.3cm},clip]{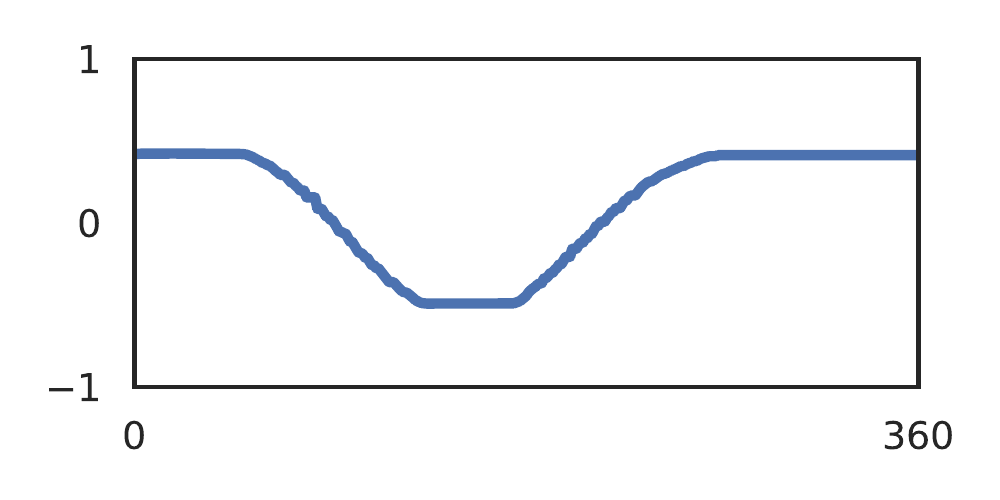}
    }
    \caption{An example pattern from the GestureMidAirD1 dataset. The original time series is shown in (a) with indicators showing the shortest and longest pattern in the training set. (b) shows the time series with zero padding, (c) is truncated to the same size as the smallest pattern in the training set, (d) uses resampling, and (e) is the proposed method.} \label{fig:example}
    \vspace{-2mm}
\end{figure}

However, one limitation of batch and mini-batch training is that the input patterns should be the same size. 
While it is still possible to use varying-sized time series in some neural network models by using one sample at a time, it comes with the sacrifice of not being able to use mini-batches. 
Therefore, time series length normalization methods are typically used, such as zero padding or truncation. 

The issue is that zero padding and truncation are usually applied without consideration of the underlying time series and do not always produce realistic patterns, as shown in Fig.~\ref{fig:example}. 
Therefore, we propose Nearest Guided Warping~(NGW), a method of normalizing the length of time series in a dataset. 
Guided warping, originally developed for data augmentation, generates data by warping the time steps of one time series to the time steps of a reference time series~\cite{iwana2021time}. 
It does this by exploiting DTW for its dynamic alignment property.
We adapt this idea to time series length normalization. 
By warping the time series patterns in a dataset to fixed-size reference patterns, the lengths of the time series can be normalized.

The contributions of this paper are as follows:
\begin{itemize}
    \item We propose a new method of normalizing the length of time series for training ANNs with mini-batches. The method warps time series to fixed lengths using DTW.
    \item A thorough investigation on time series length normalization methods is performed. We test 17 time series length normalization methods, an experiment without batches, and the proposed method on the 11 datasets in the UCR Time Series Archive~\cite{UCRArchive2018} that have pattern lengths that vary.
    \item Through the experiments, we show that the way that fixing the varying length time series can have a dramatic effect on the accuracy. 
    \item Analysis and ablation are performed to demonstrate the effectiveness of the proposed method.
\end{itemize}

\section{Related Work}
\label{sec:related}

While many works use a form of time series length normalization, such as padding or truncation, very few works compare the effects that it has on the model. 
Dwarampudi and Reddy~\cite{dwarampudi2019effects} compared the use of zero padding on an LSTM and CNN. 
In this work, the authors performed pre-padding and post-padding and compared the accuracies on a Natural Language Processing (NLP) dataset.
Lopez-del Rio et al.~\cite{Lopez_del_Rio_2020} compared multiple padding methods in tackling protein classification. 
They pad with zeros using pre-padding and post-padding, as well as propose mid-padding, ext-padding, rnd-padding, aug-padding, strf-padding, and zoom-padding.

There have also been previous works that use time series length normalization for classical methods. 
For example, uniform scaling is often used in time series motif discovery~\cite{Keogh_2003,Yankov_2007}.
Another example is Tan et al.~\cite{tan2019time} compared time series normalization methods with using a nearest neighbor classifier.

\section{Nearest Guided Warping}

The purpose of input length normalization is to fix the number of time steps in time series data to be the same so that it can be used with batch and mini-batch training with ANNs. 
Given sequence $\mathbf{x}=x_1,\dots,x_t,\dots,x_T$, where $x_t$ can be univariate or multivariate, we want to ensure length $T$ is the same for all time series in dataset $\mathbf{x}\in\mathbf{X}$. 
In order to do this, we exploit the time warping ability of DTW to stretch shorter patterns to be the same length as the longer patterns while maintaining the local features.

\subsection{Dynamic Time Warping}
\label{sec:dtw}

Traditionally, DTW~\cite{sakoe1978dynamic} is used as a global distance measure between two time series. 
Namely, given two time series, a prototype $\mathbf{p}=p_1,\dots,p_i,\dots,p_I$ and sample $\mathbf{s}=s_1,\dots,s_j,\dots,s_J$ where $I$ and $J$ are time series lengths and $p_i$ and $s_j$ are elements at time step $i$ and $j$, DTW finds the optimal alignment between the elements of $\mathbf{p}$ and $\mathbf{s}$. 
Using the alignment, the local distances between the matched elements are combined to form a global distance measure. 

Specifically, DTW finds the minimal path on an element-wise cost matrix $C$. 
The minimal path on $C$ corresponds to the minimal distance between matching elements in $\mathbf{p}$ and $\mathbf{s}$ given constraints. 
To find the minimal path, dynamic programming is employed. 
Namely, a cumulative sum matrix $D$ is calculated using the recurrent function:
\begin{multline}
\label{eq:slope}
{D}(i, j) =
\\ {C}(p_i, s_j) + \min_{(i', j') \in \{(i-1, j), (i-1, j-1), (i-1, j-2)\}}{D}(i', j'),
\end{multline}
where ${D}(i,j)$ is the cumulative sum of the $i$-th and $j$-th elements and ${C}(p_i, s_j)$ is the local distance between $p_i$ and $s_j$. 
Elements $i'$ and $j'$ are pair of time steps for matched elements $p_{i'}$ and $s_{j'}$. 
For the experiments, Euclidean distance is used for ${C}(p_i, s_j)$.

Also, Eq.~\eqref{eq:slope} uses an asymmetric slope constraint with no window constraint. 
For the proposed method, this specific constraint is used because it ensures that the resulting number of matching always equals the number of time steps in the prototype. 
This fact is important for the proposed time series length normalization. 

\subsection{Input Length Normalization Using Guided Warping}
\label{sec:ngw}

\begin{figure}[t]
    \centering
    \includegraphics[width=0.45\columnwidth,trim={0cm 0cm 0cm 0cm},clip]{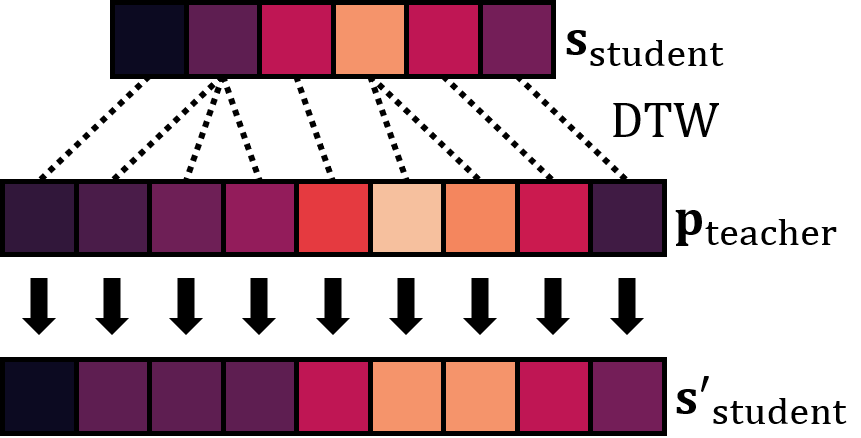}
    \caption{Illustration of guided warping. DTW is used to align the time steps of student $\mathbf{s}_\mathrm{student}$ to the time steps of teacher $\mathbf{p}_\mathrm{teacher}$ to form $\mathbf{s}'_\mathrm{student}$.} \label{fig:ngw}
\end{figure}

Provided a training set $\mathbf{X}_{\mathrm{train}}$ and test set $\mathbf{X}_{\mathrm{test}}$, each with varying number of time steps, we construct sets $\mathbf{X'}_{\mathrm{train}}$ and $\mathbf{X'}_{\mathrm{test}}$ with a fixed number of time steps. 
Note, a validation set can also be used; the process of input length normalization is exactly the same as the test set. 
As shown in Fig.~\ref{fig:ngw}, To normalize the lengths of the time series, we elongate the shorter patterns by warping in the time dimension so that they have a fixed number time steps. 
In order to do this, we use a method inspired by guided warping~\cite{iwana2021time}. 
In guided warping, time series are generated by warping a student pattern based on a teacher pattern for the purpose of generating time series data augmentation. 
This is done by aligning the elements in the two time series using DTW and warping the student pattern by the teacher pattern. 
Thus, the generated pattern has the features of the student pattern set to the time steps of the teacher pattern. 
This is opposed to linearly warping in interpolation and Zoom padding~\cite{Lopez_del_Rio_2020}, or randomly warping~\cite{le2016data,Um_2017}.

Namely, by tracing the minimal route from $D(I,J)$ to $D(0,0)$ on the cumulative sum matrix $D$ determined by Eq.~\eqref{eq:slope}, a matching between the elements at $i'$ and $j'$ is formed.
As mentioned previously, the advantage of using the asymmetric slope constraint, the number of matches between $i'$ and $j'$ always equals the number of time steps in the prototype. 
Therefore, if the number of time steps in the prototype is fixed, then the number of matches will be fixed. 
Thus, by warping using DTW with fixed-size prototypes, fixed size patterns can be generated using guided warping.

\begin{figure}[t]
    \centering
    \includegraphics[width=1\columnwidth,trim={0cm 0cm 0cm 0cm},clip]{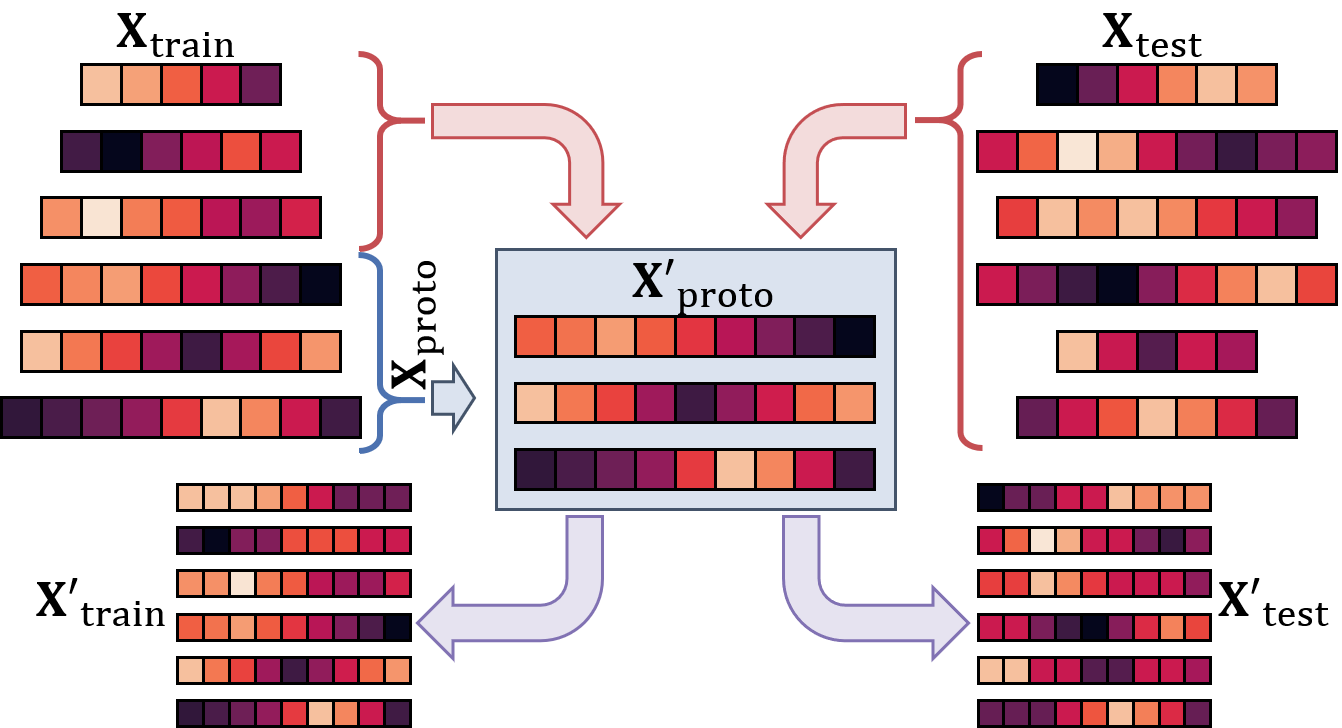}
    \caption{Diagram of the proposed method of fixing the size of time series with varying length.} \label{fig:process}
    \vspace{-2mm}
\end{figure}

In the previous use of guided warping, the reference pattern can be selected randomly or using a discriminator~\cite{iwana2021time}. 
Instead, we propose Nearest Guide Warping~(NGW), shown in Fig.~\ref{fig:process}. 
In NGW, a prototype set $\mathbf{X}_\mathrm{proto}$ is created from patterns selected from training set $\mathbf{X}_{\mathrm{train}}$. 
To select the time series for $\mathbf{X}_\mathrm{proto}$, the top $\alpha$ quantile of lengths are used. 
The selected time series in $\mathbf{X}_\mathrm{proto}$ are then resampled so that they have the same length $\mathbf{X}'_\mathrm{proto}$. 
Next, the nearest $\mathbf{X}'_\mathrm{proto}$ to each time series $\mathbf{s}_\mathrm{student}$ in $\mathbf{X}_\mathrm{test}$ or $\mathbf{X}_\mathrm{train}|[\mathbf{X}_\mathrm{train}\neq\mathbf{X}_\mathrm{proto}]$ is found, or:
\begin{equation}
\label{eq:distance}
\mathbf{p}_\mathrm{teacher}=\underset{\mathbf{p}\in\mathbf{X}'_\mathrm{proto}}{\mathrm{arg\,min}} \ d(\mathbf{s}_\mathrm{student},\mathbf{p}),
\end{equation}
where $d(\mathbf{s}_\mathrm{student},\mathbf{p})$ is the DTW distance between $\mathbf{s}_\mathrm{student}$ and $\mathbf{p}$.
Finally, the time steps of $\mathbf{s}_\mathrm{student}$ are set to the time steps of the nearest prototype $\mathbf{p}_\mathrm{teacher}$. 
The result is time series $\mathbf{s}'_\mathrm{student}$ with a fixed number of time steps.

As mentioned previously, the quantile threshold $\alpha$ determines which time series in $\mathbf{X}_\mathrm{train}$ are used for $\mathbf{X}_\mathrm{proto}$. 
Then, $\mathbf{X}_\mathrm{proto}$ are resampled to be the same length to form $\mathbf{X}'_\mathrm{proto}$. 
However, the target length does not necessarily need to be the maximum length. 
Instead, a second quantile $\beta$ can be used to find the target length. 
The value of $\beta$ should be within $[\alpha, \max_{\mathbf{x}\in\mathbf{X}}{T}]$.
The parameters for $\alpha$ and $\beta$ are hyperparameters that need to be defined.

\section{Experimental Results}
\label{sec:experiment}

\subsection{Datasets}

In order to evaluate the proposed methods, datasets from the 2018 UCR Time Series Archive~\cite{UCRArchive2018} are used. 
The datasets, listed in Table~\ref{tab:properties}, used in the experiments are all 11 datasets that have varying lengths. 
As seen in the table, the lengths of each time series can vary greatly, with PLAID having the largest discrepancy of 100 time steps as the shortest and 1,344 time steps as the longest.

\begin{table}[!t]
    \caption{Comparison of the Lengths of the Time Series in Each of the Training Datasets}
    \label{tab:properties}
    \centering
    \begin{tabular}{lcccc}
    \hline
    Dataset & Type &  Min. & Max. & Ave. \\ \hline
    AllGestureWiimoteX & Sensor & 11 & 385& 124.9 \\
    AllGestureWiimoteY & Sensor & 8 & 369 & 128.6\\
    AllGestureWiimoteZ & Sensor & 33 & 326 & 125.5\\
    GestureMidAirD1 & Trajectory & 80 & 360  & 166.5\\
    GestureMidAirD2 & Trajectory & 80 & 360 & 166.5 \\
    GestureMidAirD3 & Trajectory & 80 & 360 & 166.5 \\
    GesturePebbleZ1 & Sensor & 115 & 455 & 233.7 \\
    GesturePebbleZ2 & Sensor & 100 & 455 & 223.5 \\
    PickupGestureWiiZ & Sensor & 29 & 361 & 145.9 \\
    PLAID & Device & 100 & 1344 & 323.8 \\
    ShakeGestureWiiZ & Sensor & 41 & 385 & 171.9 \\
    \hline
    \end{tabular}
    \vspace{-2mm}
\end{table}

For the experiments, the preset training and test sets are used. 
As for preprocessing, the features are min-max normalized so that the values of the elements in the training set are between [-1, 1] for each of the respective datasets. 
After the min-max normalization, one of the various time series length normalization methods is used.

\subsection{Experimental Settings}

\subsubsection{Neural Network Models}

For the experiments, three neural networks were used: a temporal CNN and two RNNs. 
The CNN is an adaptation of a Visual Geometry Group network (VGG)~\cite{simonyan2014very} for time series (1D VGG) with the same hyperparameters as the original VGG, except that it uses 1D convolutions. 
However, due to the difference in sequence size compared to the original VGG, we use a variable number of blocks as outlined in Iwana and Uchida~\cite{iwana2021an} in order to avoid excessive pooling.
Each block uses 1D convolutions of filter length 3 that were initialized using a uniform variance~\cite{He_2015_ICCV}. 
Following the convolutional blocks, there are two fully connected layers with 4,096 nodes, rectified linear units (ReLU), and dropout with a probability of 0.5. 
We follow the same training procedure as the original VGG using batch size 256 with gradient descent and an initial learning rate of 0.01 with weight decay of $5\times 10^{-4}$ and momentum of 0.9.

For the RNNs, we use a Long-Short Term Memory RNN (LSTM)~\cite{Hochreiter_1997} and a Bidirectional LSTM (BLSTM)~\cite{Schuster_1997}.
Both are used in order to demonstrate the differences between pre, post, and outer padding.
The hyperparameters for the LSTM and BLSTM were taken from the suggestions determined by Reimers and Gurevych~\cite{reimers2017optimal}. 
Specifically, the LSTM has one layer with 100 units and the BLSTM has two layers with 100 units. 
Both are trained using batch size 32 and Nesterov Momentum Adam~(Nadam)~\cite{dozat2016incorporating} with an initial learning rate of 0.001 as suggested by Reimers and Gurevych. 

All of the networks were trained for 20,000 iterations. 
In addition, for all of the experiments, we trained each network, dataset, and length normalization method five times and used the median value. 
This is to ensure the reliability of the results.

\subsubsection{Input Length Normalization Methods}

In order to evaluate the proposed method, we used a variety of time series length normalization methods, including standard practice methods and methods from literature. 
The following methods are compared:
\begin{itemize}
    \item \textbf{Zero Pad} is the most common method. It pads the shorter time series with zeros to make them the same length as the longest pattern in the dataset. The idea is that zeros should be ignored by the weights. Zero Pad (Pre) and Zero Pad (Post) pad the time series before and after, respectively. We also implement the Zero Pad (Outer) and Zero Pad (Mid) variations as proposed by Lopez del Rio et al.~\cite{Lopez_del_Rio_2020}. 
    \item \textbf{Truncate} is a well-established method that truncates the longer time series to the shortest time series in the training set. The variations, Truncate (Pre), (Post), and (Outer), are used. It should be noted that there is a rare chance that a time series in the test set is shorter than any in the training set. In this case, we zero pad the short test set pattern until it is the same length as the training set.
    \item \textbf{Noise Pad}, proposed by Tan et al.~\cite{tan2019time}, is similar to Zero Pad, but instead of zeros, low amplitude noise is used. As in Tan et al., the noise is generated from a uniform distribution within [0, 0.001]. Also, Noise Pad (Pre), (Post), and (Outer) are compared.
    \item \textbf{Edge Pad} is also similar to Zero Pad. Edge Pad (Pre) pads with the first element of the time series. Edge Pad (Post) pads with the last element. Edge Pad (Outer) pads with both. 
    \item \textbf{Resampling} uses linear interpolation to stretch all of the time series in the dataset to be the same length as the largest pattern.
    \item \textbf{STRP Pad} and \textbf{Random Pad} were proposed by Lopez del Rio et al.~\cite{Lopez_del_Rio_2020}. They are similar in that zeros are injected throughout the time series. For STRP Pad, the zeros are distributed evenly in the time series and for Random Pad, the zeros are added randomly.
    \item \textbf{Zoom Pad}~\cite{Lopez_del_Rio_2020} is similar to Resampling, but instead of interpolating between elements, time steps are repeated. This could be considered similar to the proposed method, except that the time series are linearly warped instead of guided by DTW.
    \item \textbf{None} does not use any length normalization method. To do this, patterns must be given to the network one at a time (i.e., batch size $B=1$). For RNNs, this is a natural way to handle time series of varying lengths. For the 1D VGG, the network is modified using Global Average Pooling (GAP) instead of flattening before the fully connected layers in order to be able to use variable-sized inputs.
    \item \textbf{NGW-$\alpha$ (Proposed)} is the proposed method with $\alpha=0.4$. The time series are warped to the size of the longest time series in the dataset. 
    \item \textbf{NGW-$\alpha$ CW (Proposed)} is a class-wise version of the proposed method that uses the top $\alpha$ longest prototypes of each class separately. 
    \item \textbf{NGW-$\alpha\beta$ (Proposed)} uses $\alpha=0.4$ and $\beta=0.7$. 
    \item \textbf{NGW-$\alpha\beta$ CW (Proposed)} is the same as NGW-$\alpha\beta$ but using a class-wise $\alpha$. 
\end{itemize}

\subsection{Results}

\begin{table}[!t]
    \caption{Comparison of Length Normalization Methods. Mean Test Accuracy (\%) of the 11 Datasets Using the Median of Five Trained Models}
    \label{tab:results}
    \centering
    \begin{tabular}{lccc}
    \hline
     & \multicolumn{3}{c}{Model} \\
    Method & 1D VGG & LSTM & BLSTM \\ \hline
    Zero Pad (Pre) & 63.61 & 46.36 & 55.16 \\
    Zero Pad (Post) & 66.12 & 21.97 & 44.35 \\
    Zero Pad (Outer)~\cite{Lopez_del_Rio_2020} & 66.57 & 17.26 & 14.46 \\
    Zero Pad (Mid)~\cite{Lopez_del_Rio_2020} & 64.49 & 43.54 & {56.65} \\
    Edge Pad (Pre) & 61.49 & 35.36 & 51.48 \\
    Edge Pad (Post) & 63.61 & 27.05 & 37.84 \\
    Edge Pad (Outer) & 63.88 & 28.37 & 31.50 \\
    Noise Pad (Pre) & 63.35 & 46.28 & 52.10 \\
    Noise Pad (Post)~\cite{tan2019time} & {66.80} & 19.07 & 44.43 \\
    Noise Pad (Outer)~\cite{tan2019time} & \textbf{67.83} & 18.20 & 24.04 \\
    Truncate (Pre) & 38.77 & 35.83 & 37.54 \\
    Truncate (Post) & 35.75 & 30.49 & 32.37 \\
    Truncate (Outer) & 38.65 & 37.50 & 39.27 \\
    Resampling & 59.26 & 40.67 & 48.55 \\
    STRP Pad~\cite{Lopez_del_Rio_2020} & 65.90 & 41.91 & 45.18 \\
    Random Pad~\cite{Lopez_del_Rio_2020} & 57.84 & 42.13 & 41.23 \\
    Zoom Pad~\cite{Lopez_del_Rio_2020} & 61.44 & 44.04 & 50.51 \\
    None ($B=1$)  & 38.23 & 31.92 & 40.55 \\
    NGW-$\alpha$ (Proposed) & 56.30 & 39.88 & 50.87 \\
    NGW-$\alpha$ CW (Proposed) & 56.30 & 41.14 & 49.07 \\
    NGW-${\alpha\beta}$ (Proposed) & 63.86 & 50.95 & 56.21 \\
    NGW-${\alpha\beta}$ CW (Proposed) & 59.11 & \textbf{53.74} & \textbf{57.62}\\
    \hline
    \end{tabular}
    \vspace{-2mm}
\end{table}

The results are shown in Table~\ref{tab:results}.
As the table shows, how the lengths of the time series are normalized has a dramatic effect on the accuracy of the accuracy. 

In general, for the tested datasets, Truncate performed poorly for all of the models. 
This is due to Truncate removing too much information. 
As shown in Table~\ref{tab:properties}, there is a large difference between the minimum time series lengths and the average. 
For some datasets like AllGestureWIimoteY, Truncate removes all but a few time steps.

Conversely, the methods that stretch the time series, such as Resampling, STRP Pad, Random Pad, Zoom Pad, and the proposed NGW, performed generally well. 
On the 1D VGG, in particular, they performed similarly. 
This is likely due to maxpooling. 
The result is that these methods would result in similar high-level features in the fully connected layers. 

The padding-based methods, on the other hand, have interesting results. 
Similar to 1D VGG, the padding method and location did not affect the results significantly. 
However, unlike 1D VGG, the location of the padding had dramatic effects on the accuracy of the RNN-based models.
For example, similar to the findings in~\cite{dwarampudi2019effects}, Zero Pad (Pre) has more than double the accuracy as Zero Pad (Post) for LSTM. 
This is because, in Zero Pad (Post), the final time steps are often zeros which affect the output of the embedding of the LSTM for softmax.
The same effect can be found in BLSTM with Zero Pad (Outer) versus Zero Pad (Mid) for a similar reason.

As for the proposed method, NGW-$\alpha\beta$ CW performed the best. 
Notably, it had the best results on LSTM and BLSTM compared to any other method. 
For the 1D VGG, the results were competitive to the comparison methods, but they were lower than some Zero Pad and Noise Pad (Post) and (Outer). 
This may be because low information padding will be removed from maxpooling.

\subsection{Effects of $\alpha$ and $\beta$}
\begin{figure}[t]
    \centering
    \subfigure[Different $\alpha$]{
    \label{subfig:alpha}
    \includegraphics[width=0.475\columnwidth,trim={0.5cm 0.5cm 0.4cm 0.3cm},clip]{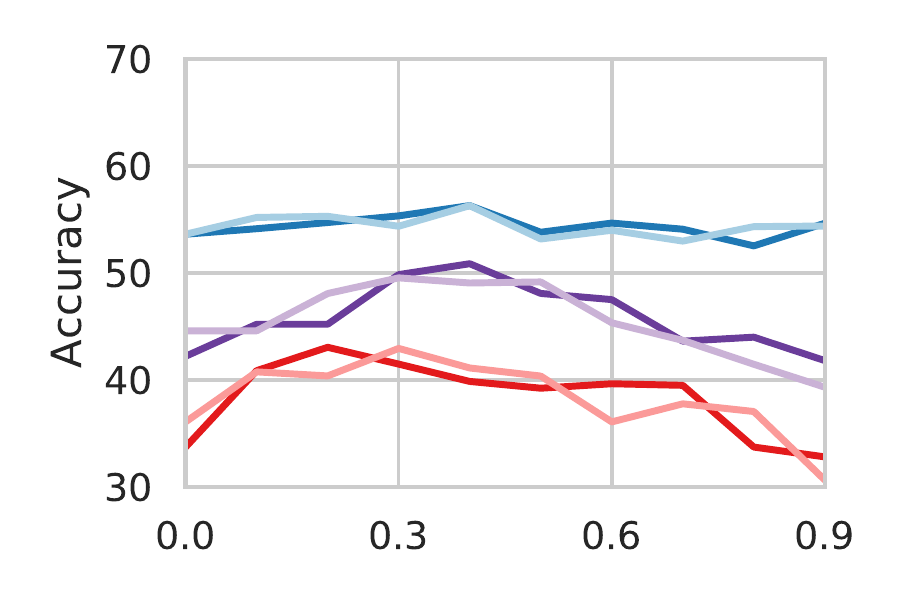}
    }
    \subfigure[Different $\beta$]{
    \label{subfig:beta}
    \includegraphics[width=0.475\columnwidth,trim={0.5cm 0.5cm 0.4cm 0.3cm},clip]{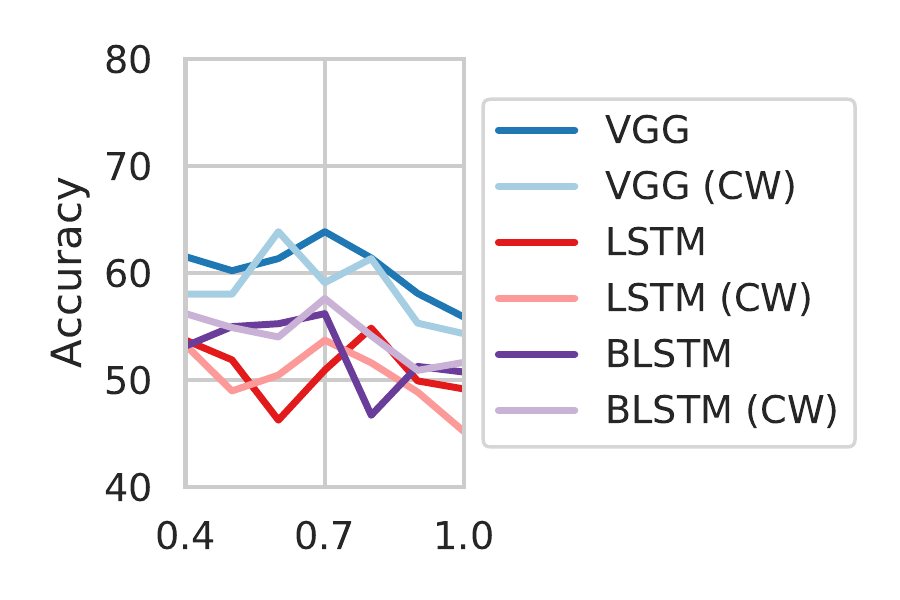}
    }
    \caption{The effects of changing $\alpha$ and $\beta$. For (a), $\beta=1$. For (b), $\alpha=0.4$.} \label{fig:alphabeta}
    \vspace{-3mm}
\end{figure}

The proposed method uses hyperparameters $\alpha$ and $\beta$. 
Fig.~\ref{fig:alphabeta} shows the effects of increasing and reducing $\alpha$ and $\beta$. 
A higher $\alpha$ means that fewer prototypes are used. 
However, a lower $\alpha$ means that the prototypes are stretched more. 
Thus, $\alpha=0.4$ is used for the experiments. 
Generally, a $\beta$ that is near the midpoint of $\alpha$ and $T$ is the best. 
This gives the best speed without excessive warping.

\subsection{Limitations}
While the proposed method has good results, there are some limitations. 
For instance, the other input length normalization methods are executed in $O(T)$ time for each time series, where $T$ is the length. 
One limitation of the proposed method is that it has a $O(T^2P)$ complexity, where $P$ is the number of prototypes due to DTW. 
Another limitation is that the asymmetric slope constraint in Eq.~\eqref{eq:slope} requires prototype $\mathbf{p}$ to be at least 0.5 times the length of $\mathbf{s}_\mathrm{student}$. 
For low values of $\alpha$, it is possible that the $\mathbf{p}$ is too short. 
Therefore in our implementation, if $\mathbf{s}_\mathrm{student}$ is longer than 2 times the length of $\mathbf{p}$, we resample it before NGW is applied.

\section{Conclusion}
\label{sec:conclusion}

In this paper, we propose a method of normalizing the input length of time series. 
The proposed method exploits DTW to warp the shorter time series by the time steps of fixed-sized time series.
Through this, the features of the short time series are dynamically aligned to common features of the fixed-sized time series. 

In addition, we perform an extensive review of input length normalization methods for time series by testing 17 methods, the proposed method, and one with no mini-batch training on a CNN and two RNNs. 
We found that something as fundamental as normalizing the input lengths can have a significant effect on the accuracy of the network.

\bibliographystyle{IEEEbib}
\bibliography{vary}

\begin{thebibliography}{10}

\bibitem{tan2019time}
Chang~Wei Tan, Fran{\c{c}}ois Petitjean, Eamonn Keogh, and Geoffrey~I Webb,
\newblock ``Time series classification for varying length series,''
\newblock {\em arXiv preprint arXiv:1910.04341}, 2019.

\bibitem{ding2008querying}
Hui Ding, Goce Trajcevski, Peter Scheuermann, Xiaoyue Wang, and Eamonn Keogh,
\newblock ``Querying and mining of time series data,''
\newblock {\em Proc. Very Larg. Data Base Endow.}, vol. 1, no. 2, pp.
  1542--1552, 2008.

\bibitem{Leng_2008}
Mingwei Leng, Xiaoyun Chen, and Longjie Li,
\newblock ``Variable length methods for detecting anomaly patterns in time
  series,''
\newblock in {\em ISCID}, 2008.

\bibitem{Jalalian_2013}
Arash Jalalian and Stephan~K. Chalup,
\newblock ``{GDTW}-p-{SVMs}: Variable-length time series analysis using support
  vector machines,''
\newblock {\em Neurocomputing}, vol. 99, pp. 270--282, jan 2013.

\bibitem{sakoe1978dynamic}
Hiroaki Sakoe and Seibi Chiba,
\newblock ``Dynamic programming algorithm optimization for spoken word
  recognition,''
\newblock {\em IEEE Trans. Acoustics, Speech, and Sig. Process.}, vol. 26, no.
  1, pp. 43--49, 1978.

\bibitem{Kahveci}
T.~Kahveci and A.~Singh,
\newblock ``Variable length queries for time series data,''
\newblock in {\em ICDE}, 2002.

\bibitem{Chandrakala_2010}
S.~Chandrakala and C.~Chandra Sekhar,
\newblock ``Classification of varying length time series using example-specific
  adapted gaussian mixture models and support vector machines,''
\newblock in {\em SPCOM}, 2010.

\bibitem{Lecun_1998}
Y.~Lecun, L.~Bottou, Y.~Bengio, and P.~Haffner,
\newblock ``Gradient-based learning applied to document recognition,''
\newblock {\em Proc. {IEEE}}, vol. 86, no. 11, pp. 2278--2324, 1998.

\bibitem{rumelhart1988learning}
David~E Rumelhart, Geoffrey~E Hinton, and Ronald~J Williams,
\newblock ``Learning representations by back-propagating errors,''
\newblock {\em Nat.}, vol. 323, no. 6088, pp. 533--536, 1986.

\bibitem{Wang_2017}
Zhiguang Wang, Weizhong Yan, and Tim Oates,
\newblock ``Time series classification from scratch with deep neural networks:
  A strong baseline,''
\newblock in {\em IJCNN}, 2017.

\bibitem{bai2018empirical}
Shaojie Bai, J~Zico Kolter, and Vladlen Koltun,
\newblock ``An empirical evaluation of generic convolutional and recurrent
  networks for sequence modeling,''
\newblock {\em arXiv preprint arXiv:1803.01271}, 2018.

\bibitem{ruder2016overview}
Sebastian Ruder,
\newblock ``An overview of gradient descent optimization algorithms,''
\newblock {\em arXiv preprint arXiv:1609.04747}, 2016.

\bibitem{iwana2021time}
Brian~Kenji Iwana and Seiichi Uchida,
\newblock ``Time series data augmentation for neural networks by time warping
  with a discriminative teacher,''
\newblock in {\em ICPR}, 2021.

\bibitem{UCRArchive2018}
Hoang~Anh Dau, Eamonn Keogh, Kaveh Kamgar, Chin-Chia~Michael Yeh, Yan Zhu,
  Shaghayegh Gharghabi, Chotirat~Ann Ratanamahatana, Yanping, Bing Hu, Nurjahan
  Begum, Anthony Bagnall, Abdullah Mueen, Gustavo Batista, and Hexagon-ML,
\newblock ``The ucr time series classification archive,'' 2018,
\newblock https://www.cs.ucr.edu/~eamonn/time\_series\_data\_2018/.

\bibitem{dwarampudi2019effects}
Mahidhar Dwarampudi and NV~Reddy,
\newblock ``Effects of padding on lstms and cnns,''
\newblock {\em arXiv preprint arXiv:1903.07288}, 2019.

\bibitem{Lopez_del_Rio_2020}
Angela~Lopez del Rio, Maria Martin, Alexandre Perera-Lluna, and Rabie Saidi,
\newblock ``Effect of sequence padding on the performance of deep learning
  models in archaeal protein functional prediction,''
\newblock {\em Scientific Reports}, vol. 10, no. 1, 2020.

\bibitem{Keogh_2003}
Eamonn Keogh,
\newblock ``Efficiently finding arbitrarily scaled patterns in massive time
  series databases,''
\newblock in {\em PKDD}, 2003, pp. 253--265.

\bibitem{Yankov_2007}
Dragomir Yankov, Eamonn Keogh, Jose Medina, Bill Chiu, and Victor Zordan,
\newblock ``Detecting time series motifs under uniform scaling,''
\newblock in {\em ACM SIGKDD}, 2007.

\bibitem{le2016data}
Arthur {Le Guennec}, Simon Malinowski, and Romain Tavenard,
\newblock ``Data augmentation for time series classification using
  convolutional neural networks,''
\newblock in {\em IWAATD}, 2016.

\bibitem{Um_2017}
Terry~T. Um, Franz M.~J. Pfister, Daniel Pichler, Satoshi Endo, Muriel Lang,
  Sandra Hirche, Urban Fietzek, and Dana Kuli{\'{c}},
\newblock ``Data augmentation of wearable sensor data for parkinson's disease
  monitoring using convolutional neural networks,''
\newblock in {\em ACM ICMI}, 2017, pp. 216--220.

\bibitem{simonyan2014very}
Karen Simonyan and Andrew Zisserman,
\newblock ``Very deep convolutional networks for large-scale image
  recognition,''
\newblock {\em arXiv preprint arXiv:1409.1556}, 2014.

\bibitem{iwana2021an}
Brian~Kenji Iwana and Seiichi Uchida,
\newblock ``An empirical survey of data augmentation for time series
  classification with neural networks,''
\newblock {\em PLOS ONE}, 2021.

\bibitem{He_2015_ICCV}
Kaiming He, Xiangyu Zhang, Shaoqing Ren, and Jian Sun,
\newblock ``Delving deep into rectifiers: Surpassing human-level performance on
  imagenet classification,''
\newblock in {\em IEEE ICCV}, 2015.

\bibitem{Hochreiter_1997}
Sepp Hochreiter and J{\"u}rgen Schmidhuber,
\newblock ``Long short-term memory,''
\newblock {\em Neural Computation}, vol. 9, no. 8, pp. 1735--1780, 1997.

\bibitem{Schuster_1997}
M.~Schuster and K.K. Paliwal,
\newblock ``Bidirectional recurrent neural networks,''
\newblock {\em {IEEE} Trans. Sig. Process.}, vol. 45, no. 11, pp. 2673--2681,
  1997.

\bibitem{reimers2017optimal}
Nils Reimers and Iryna Gurevych,
\newblock ``Optimal hyperparameters for deep lstm-networks for sequence
  labeling tasks,''
\newblock {\em arXiv preprint arXiv:1707.06799}, 2017.

\bibitem{dozat2016incorporating}
Timothy Dozat,
\newblock ``Incorporating nesterov momentum into adam,''
\newblock in {\em ICLR Workshops}, 2016.

\end{thebibliography}

\end{document}